# Evaluation of Postural Muscle Synergies during a Complex Motor Task in a Virtual Reality Environment

Xupeng Ai[1], Victor Santamaria[1], Isirame Babajide Omofuma[1], and Sunil K. Agrawal[1,2*]

[1]Xupeng Ai, [1]Victor F. Santamaria and [1]Isirame Babajide Omofuma are with the Department of Mechanical Engineering, Columbia University, New York, NY 10027 USA.

[1,2*]Sunil K. Agrawal is with the Department of Rehabilitation and Regenerative Medicine, Columbia University, New York, NY 10027 USA, and also with the Department of Mechanical Engineering, Columbia University, New York, NY 10027 USA (e-mail: sa3077@columbia.edu).

*Abstract*—In this study, we investigate how the central nervous system (CNS) organizes postural control synergies when individuals perform a complex catch-and-throw task in a virtual reality (VR) environment. A Robotic Upright Stand Trainer (RobUST) platform, including surface electromyography and kinematics, was used to investigate how the CNS fine-tunes postural synergies with perturbative and assist-as-needed force fields. A control group without assistive forces was recruited to elucidate the effect of force fields on motor performance and postural synergy organization after the perturbation and during the VR reaching task. We found that the application of assistive forces significantly improved reaching and balance control. The group receiving assistive forces displayed four postural control synergies characterized by higher complexity (i.e., greater number of muscles involved). However, control subjects displayed eight synergies that recruited less number of muscles. In conclusion, assistive forces reduce the number of postural synergies while increasing the complexity of muscle module composition.

*Index Terms* – virtual reality, trunk perturbations, robotic stand trainer, pelvic assistance, force fields, muscle synergy

## I. INTRODUCTION

The central nervous system (CNS) constrains muscle activations spatially and temporally to respond to self-triggered movements, e.g., during reaching or during unanticipated external perturbations [1,2]. Within a systems theory view, the neural and biomechanical control of a goal-oriented and dexterous task may be simplified by muscle synergies [3]. Some authors have proposed that the CNS might apply muscle synergies as execution-level sensorimotor solutions to transform task-level motor intentions into detailed spatiotemporal muscle patterns in order to reduce motor complexity [1]. As a result, the CNS would combine muscle synergies to cope with the intricacies of the neuronal computations at the cerebral, brainstem, and spinal cord levels as well as to meet the physical demands of the surrounding environment.[4]

In-place upright postural control, with a fixed base of support (BOS), is achieved by scaled and timely ascending and descending muscle responses around two primary joints: hip and ankle strategies [5–7]. These two postural modes of control are not mutually exclusive and can co-exist for the dual control of postural orientation and balance [8]. The successful attainment of upright stability against gravity relies on feedback (i.e., online modulation of postural responses) and feedforward strategies (i.e., postural control mechanisms, established prior to an action such as a reach).[9–11]. Several animal and human studies have shed light on the role of muscle synergies to control posture and generate direction-specific muscle forces in the lower limbs during the loss of balance [12,13], and in arm control during fast and multidirectional reaches [2,14]. However, most of the previous research on postural control have dealt with how the CNS copes with postural imbalance during unsupported standing (i.e., reactive postural control). In our present study, we address the fundamental question of how the CNS organizes postural muscle adjustments during postural imbalance and during the execution of a complex goal-oriented task. In addition, we studied whether external postural assistance modulates the organization of postural control synergies.

To address the research question—which involves the CNS organization of automatic and volitional postural responses (APR and VPR, respectively) —we investigated a perceptual serial three-step upper extremity task performed after a period of postural imbalance. The goal-oriented task was perceptual because the subjects had to first spatially localize a flying ball and aim at a target. The task was complex because the subjects had to catch a ball, aim, and then throw at a target. Additionally, we investigated if the CNS would use different postural synergies in presence of external postural assistance tailored to the subjects' base of support (BOS). [15,16] In the ROAR laboratory of Columbia University, we designed the Robotic Upright Stand Trainer (RobUST) platform [15,16] to study the human body's balance during standing while responding to force perturbations in the trunk and/or assistant force in the pelvis. The system uses lightweight cables to provide trunk perturbations and pelvic assistance and can measure human movements, ground reaction forces, and muscle responses. Research has shown that the activation of muscle synergies correlates with task-level variables such as the center of pressure displacement[17]. With the application of assistive force field, the task goal remains the same; however, the biomechanics involved in the control of postural stance (center of mass, or COM, with respect to the BOS) are different. The goal of force fields is to provide virtual postural assistance tailored to the individual's balance for reducing stability challenges. Neuromechanical models[1,2] lead us to think that assistive force fields may modulate the biomechanical solution space that the musculoskeletal system copes with during the performance of the postural task. Thus, we could expect a different muscle organization during a postural task in combination with assistive force fields.

In this paper, we also added a Virtual Reality (VR) catch and throw task when the subject is in RobUST. In our opinion, the use of a VR setup simplifies the experimental paradigm and circumvents methodological limitations that would be present in a physical environment. . We investigated a complex serial four-step catch-and-throw task that consisted of (i) visually tracking a flying ball, (ii) catching the ball with the hand remote controller, (iii) aiming at a moving target, and (iv) throwing at the target to accurately hit the target center. All participants received RobUST-mediated perturbative forces on the trunk as they were visually tracking the flying ball. Additionally, a group of subjects received tailored assistive pelvic force fields when the subject's pelvic center traversed to or beyond a stable boundary. This complex catch-and-throw task allowed us to explore whether "assist-as-needed" pelvic force field can modulate postural synergies during this task.

## II. METHODOLOGY

### A. Study & Subjects

The study protocol (AAAR6780) complied with the ethics guidelines for human subject research, as established by the Institutional Review Board of Columbia University. We confirm that all experimental protocols were approved by the Institutional Review Board of Columbia University and performed in accordance with relevant guidelines and regulations. We confirm that informed consent was obtained from all subjects or, if subjects are under 18, from a parent and/or legal guardian. The study follows an experimental design with subject randomization for the experimental (i.e. assistive-force fields) and control groups (i.e. no assistive force-field). A total of 10 healthy young subjects (5 subjects per group) were recruited (Age = 21~36; Females = 7; and Right-handed = 8; Mean height = 132.7 ± SE = 6.5cm and Mean weight = 65.7 ± SE = 9.9kg).

### B. RobUST

RobUST is a cable-driven robotic platform that can deliver horizontal forces on the subject's upper body through belts worn on the trunk and pelvis (Fig.1). Each belt has four cable connections, and each cable is connected to the motor (Maxon Motor, Switzerland) through pulleys. The subjects stand with each foot placed on a force plate to collect force and pressure data (Bertec Force Plate V1, Ohio). A motion capture system (Vicon Vero 2.2, Denver) provides marker positions placed on the belts in real-time. We use pelvic/ trunk belt's makers' average position to represent subjects' pelvic/trunk center. This device has been described in detail in previous publications [15,16]. RobUST has two operational modes: (1) Perturbation mode: the subject will be pulled to a specific direction (randomly chosen from forward, backward, left and right) by a short pulse of force which is the resultant force created by four cables on the trunk belt; (2) Assistive-force-field mode: when the subject's pelvic center moves out of the predetermined boundary, RobUST provides the subject with an assistive recovery force, (a resultant force created by four cables on pelvic belt pointing to the original pelvic center). In this study, the trunk belt was set to the perturbation mode, and the pelvic belt was set to the assistive-force-field mode of a group of subjects. Based on the presence or absence of the assistive force field, all experimental subjects were divided into two groups. Those who received the pelvic assistive force field in the experiment were called the Force-Field group (FF group). Those who did not receive the pelvic assistive force field were called the No-Force-Field group (no-FF group). The presence or absence of the assistive force field in the pelvic plane is the critical variable in this research.

Before doing the experiment, we used the perturbation mode to detect every subject's maximum threshold perturbation in each cardinal direction. Subjects were blindfolded and used proprioception and vestibular sensed data to respond to gradually increased perturbations from different directions. The starting perturbation for each direction was set at 40% of the subject's body weight. Then perturbation force was increased by 1% if the subject maintained balance until they lost balance, which means that the subject lifted one leg off the floor. The highest force perturbation in each direction was taken as the perturbation threshold for that direction. In the subsequent experiment, every subject received perturbations at the perturbation threshold.

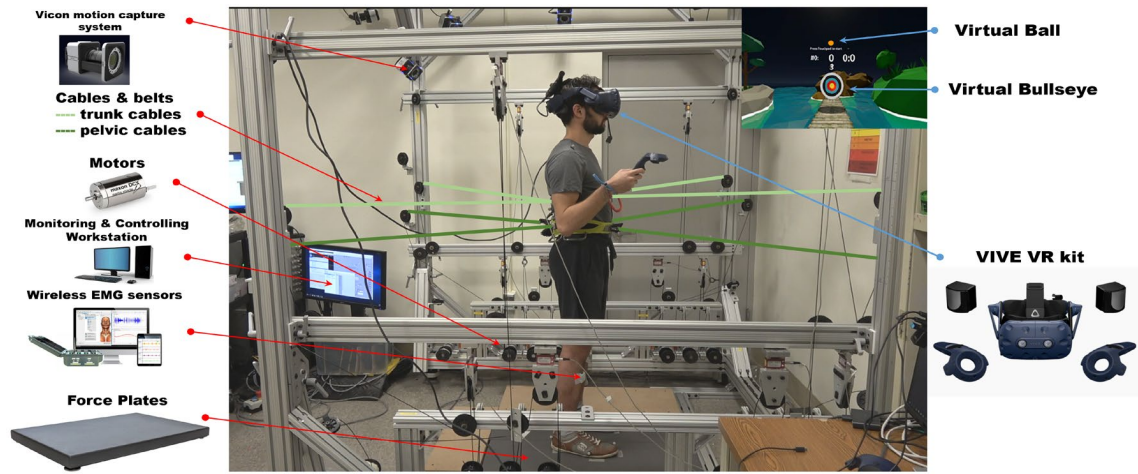

Figure 1 Illustration of experimental setup. The light green lines depict the cables attached to thorax to receive perturbative-force fields. The dark green cables represent the pelvic belt to receive assistive-force fields in the experimental group. The remote control was used to catch and throw the virtual ball.

We also experimentally determined the subject's "balance boundary" which is the union of all data points for the excursion of the pelvic center during gradually increased perturbations in which the subject did not lose balance. In the subsequent experiment, each subject in FF-group received an assistive recovery force when the pelvic center moves out of the balance boundary. Our previous paper [15,16] described how to calculate the assistive force in detail.

*C. Virtual Reality*

The subjects wore a VIVE head-mounted VR to complete a customized game that consisted of catching, aiming, and throwing a ball (Fig. 1). In the virtual environment, a subject stood on a wooden bridge surrounded by water with the bull's eye in the front. The position of the virtual bridge corresponds to the location of the force plate in the physical world. Once triggered, the ball is lodged towards the subject at the sternum height, with an uncertainty radius = 10cm. The subject had to catch the ball, aim and then throw the ball at the target within a constrained time window of 3s. Every time the subject caught and threw the virtual ball successfully within the 3s period, the motor event was recorded as successful. Otherwise, it was recorded as unsuccessful. The relative horizontal distance in the virtual world between the target and the subject was kept constant at 5m. The ball came to the user at constant speed of 30cm/s before being caught by the user. The target to throw the ball was divided into ten circular rings with the center as bullseye. For scoring, the bullseye was 10 points and the score in each outer ring decreased linearly from 10 to 0 (missed target). The VR program ensures that subjects complete the complex VR task within a 3s time window once the ball is released, during which they receive the trunk perturbation and pelvic assistance. Each subject participated in two sessions, each with 50 catch-and-throws. A break of 5min was included between sessions. (Fig.2)

*D. Data Processing*

Two force plates (Bertec Force Plate V1, Ohio) were used to collect ground reaction forces (GRF) and COP at 1,000Hz. The maximum displacement of COP along antero- posterior and lateral axes (mm) were computed during perturbations. A Surface electromyography (EMG) at 2000Hz was collected using a 14 channel Delsys Trigno Wireless System (Delsys Incorporated, Massachusetts). EMG activity was measured bilaterally from 14 leg and trunk muscles: tibialis anterior (TA), lateral gastrocnemius (LG), rectus femoris (RF), biceps femoris (BF), gluteus medius (GM), rectus abdominis (ABD), and erector spinae (ES). Force plates and EMG systems were synchronized.

A custom MATLAB program was used to filter and process EMG signals offline. EMG data was first band-passed filter (20-

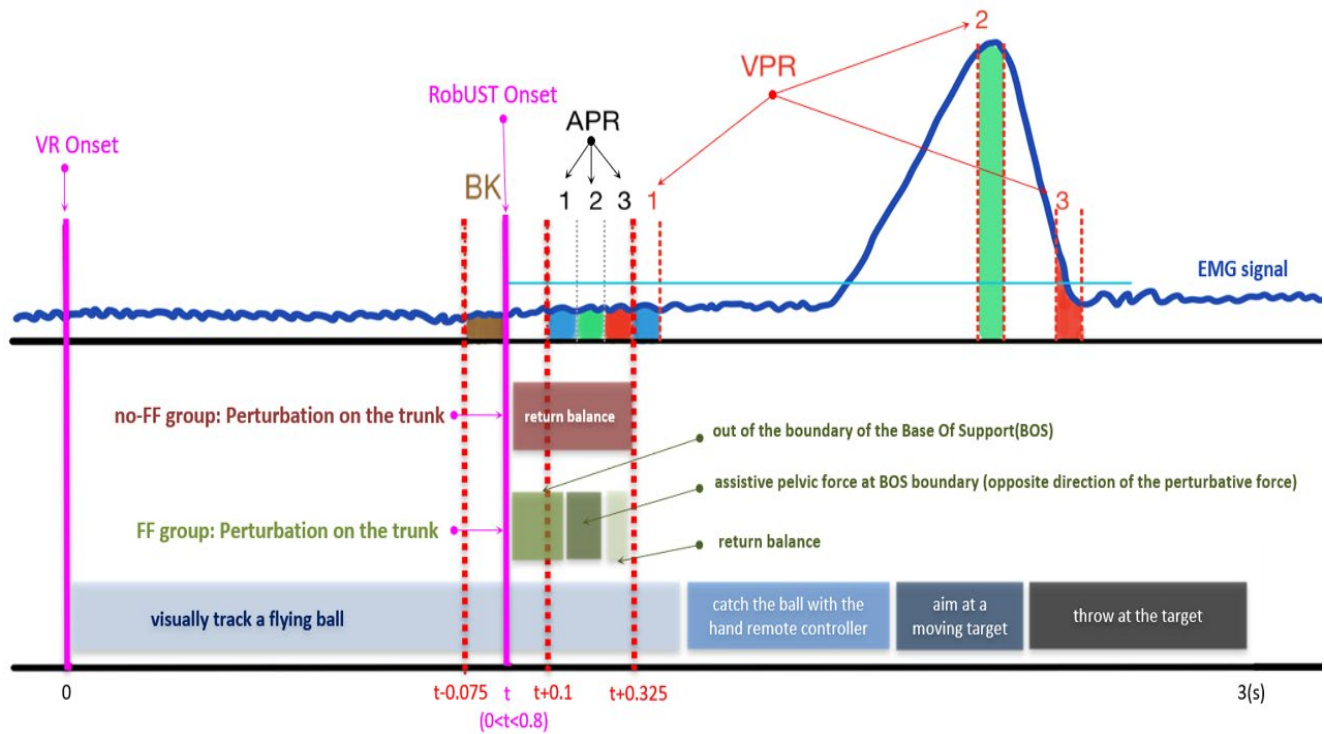

Figure 2 We define a trial as the duration when the subject completes one set of catching, aiming and throwing movements. When the trial starts, a ball flies towards the subject. As the subject is visually aiming at the flying ball(random timepoint between 0~0.8s after the ball is thrown), RobUST delivers a perturbative force in one of the four horizontal directions(forward, backward, dominant side and non-dominant side) on the torso. Before the experiment, the subjects were asked to close their eyes and receive perturbative forces on the torso that gradually increased in proportion to their body weight. Once they lost balance and took a step, we recorded this maximum perturbative force. In the VR experiment, all perturbative forces used this value for the subject. During VR task, the FF group additionally receives an assistive pelvic force field at the boundaries of the BOS that is opposite to the direction of the perturbative force. All subjects in FF and no-FF group were instructed not to move their feet unless they obligatorily had to take a step. Subjects' reaction postural control occurs in APR 1,2 & 3 bins. Their volitional control of the catch and throw movement is monitored by VPR 1,2 & 3bins.

300Hz), demeaned, rectified, and low-pass filtered at 50Hz. Following the method proposed by Torres-Oviedo et al[13,18,19], we addressed the change in EMG activity across time domains during the APR period in four time windows (or bins). In Fig.2., the pink line "VR Onset" is the timepoint at the beginning when the ball is thrown to the subject, the pink line "RobUST Onset" is the timepoint when the RobUST provides a perturbation to the subject, its time is randomly chosen in a range 0~0.8 second after the "VR Onset". We chose four bins as follows: 75ms-background (BK) before the RobUST Onset (the brown bin in Fig.2.), 100-175ms (APR1) immediately after RobUST Onset, 175-250ms (APR2) and 250-325ms (APR3) after RobUST Onset[20]. We also computed postural EMG activity while the subjects performed the complex VR catch-and-throw task to study the behavior during VPR period. Because the time periods during the catch and throw were variable across trials, perturbation directions and subjects, a fixed time window, as the one proposed for APR, was no longer suitable to accurately reflect EMG muscle activity changes during VPR. Instead, we used an amplitude threshold where we identified the maximum value of EMG activity and then used a 5% peak as threshold for muscle offsets. In Fig. 2, VPR1 would correspond to the 75ms blue bin after APR3, which is 325-400ms after RobUST Onset, VPR2 is the green bin and its center is at the peak value of EMG signal and VPR3 is the red bin whose center is at the intersection point of the light blue threshold line and the dark blue EMG signal. For all time bins (BK, APR 1,2,3 & VPR 1,2,3) in this paper, their lengths are 75ms.

A trial is seen as the whole process of catch and throw movement (Fig.2). For each of the 14 EMG muscle channels (EMG sensors are symmetrically located in subject's left and right side, 7 EMG channels in each side, 14 channels in total) in each trial, the average muscle activity in each of the seven bins was computed (BK, APR1,2,3 & VPR1,2,3). From these, we generated a data vector for each muscle that contained 14 muscle data points [7bins × 4 directions (Backwards, forwards, non-dominant side and dominant side) × 50 times catch and throw)]. For all perturbation directions and time bins across subjects, the averaged EMG data values in the vector were then normalized to their respective maximum amplitudes (iEMG) so that each muscle value was between 0 and 1 so that we can compare the averaged 7 bins' values among trials or sessions or subjects or muscles.

*E. Extraction of Muscle Synergies*

The non-negative matrix factorization method was used to extract muscle synergy from the raw EMG signals[18,21–23]. This linear decomposition method assumes that the activation state of each muscle in a given time period is composed of a linear combination of several muscle synergies. Each muscle synergy $W_i$ is activated by the synergy activation coefficient $c_i$, also called the neural command coefficient. Therefore, the net muscle activation vector M can be expressed as:

$$M = c_1 W_1 + c_2 W_2 + \ldots c_n W_n \quad (1)$$

where $W_i$ is a vector that represents the relative activation level of muscle activity in a muscle group and defined by muscle synergy $i$. Each element of $W_i$ represents the relative contribution of a muscle to the muscle synergy vector $W_i$, and the value ranges from 0 to 1. For each muscle synergy, elements in $W_i$ are constant, and the total muscle synergy is regulated by a single scalar non-negative activation coefficient $c_i$, which represents the neural command for muscle synergy. It determines the relative contribution of muscle synergy $W_i$ to the overall muscle activation pattern M. For each synergy $W_i$, the set of activations $c_i$ across all perturbation directions during BK and APRs (or VPRs) is the vector $C_i$. In the three APRs (or VPRs), $C_i$'s are the muscle tuning curves which characterize how a muscle synergy $W_i$ gets modulated by the perturbation directions and time.

$$VAF = (1 - \frac{\Sigma_{i,j}(V-V_r)^2_{i,j}}{\Sigma_{i,j} V^2_{i,j}}) \times 100\% \quad (2)$$

The number of muscle synergies is determined by the Variability Accounted For ($VAF$) test[9] defined in equation (2). $V$ is the raw EMG data matrix and $V_r$ is the reconstruction matrix. The measure guarantees that each muscle tuning curve can be well reconstructed so that the critical temporal and spatial characteristics of each muscle activation can be well explained through muscle synergy. For subjects in both groups (FF group and no-FF group), we repeated the $VAF$ calculation by changing the number of synergies (Nsyn) from 1 to 10 and then selected the least number that can reconstruct each muscle's response across all trials for all subjects. In our calculations, $VAF > 90\%$ for the complete muscle matrix which is composed of 14 muscles vectors.

*F. Statistical Analysis*

SPSS (IBM, version 27, 2020) was used for statistical analysis. The alpha rate was set at 0.05. Normal distribution of the data was examined with *Shapiro-Wilk* test and visually inspected with *Q-Q* plots. Motor behavior data (e.g., successful reaches and VR scoring) were not normally distributed and thus a non-parametric *Mann-Whitney U* was applied. COP-related kinematics between groups were examined with independent *t-tests*. In our muscle synergy analysis, a three-way mixed Analysis of Variance (*ANOVA*) with two between-factors (group: FF and no-FF; and perturbation direction: forward, backward, dominant, and non-dominant) and one repeated measures-factor (75ms-bins: 1, 2, and 3) was applied to examine significant spatial-temporal differences in muscle tuning curve data. In case of a significant ANOVA model, we proceeded with *post-hoc* testing with *Bonferroni's* inequality procedure for multiple comparisons. We studied homoscedasticity and multicollinearity of the data with *Levene's* and Mauchly's sphericity tests, respectively. In case that the assumptions of homogeneity and sphericity were violated, adjusted *p-values* to variance-corrected and Greenhouse-Geisser correction were reported.

## III. RESULTS

### A. Motor Behavior Success during the VR catch-and-throw task

Overall, subjects in the FF group performed more effectively than those in the no-FF group during the VR catch-and-throw task (Fig. 3). Subjects in the FF group showed more successful catches (Median = 50, IQR = 3) than those in the no-FF group (Median = 48, IQR =14); $U = 6.5$, $p = 0.03$, $\eta^2 = 53\%$. Similarly, subjects in the FF group were capable of executing greater number of throws (Median = 36, IQR = 20) than those in the no-FF group (Median = 26, IQR = 26); $U = 7.0$, $p = 0.04$; $\eta^2 = 51\%$. The use of pelvic assistive fore fields improved accuracy (FF group Median = 176, IQR =194 and no-FF group Median = 113, IQR = 166); however, this difference was not statistically different between groups ($U = 9.0$, $p = 0.09$).

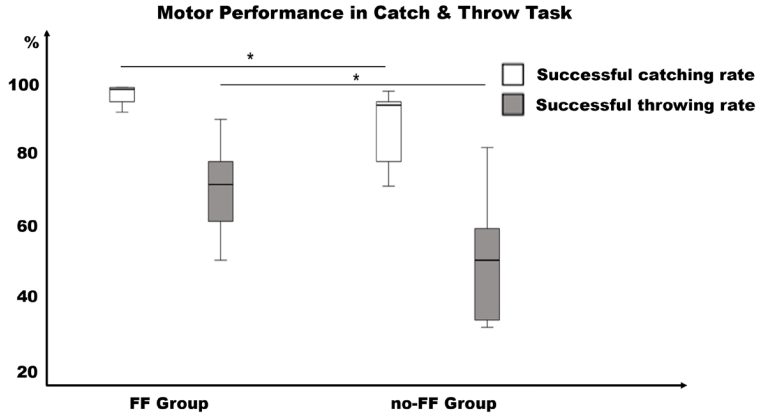

Figure 3 Motor performance during the catching and throwing components of the complex VR action. Y-axis is the successful catching/throwing rate, which is the number of successful catching/throwing trials divided by the total number of trials. The application of the assistive pelvic force field improved the success of the motor behaviors. * = p < 0.05.

### B. Balance control during postural imbalance

The subjects in the FF group performed the VR catch-and-throw task with greater level of balance control, as determined by COP-related variables (Table 1). Compared to participants in no-FF group, the FF group displayed reduced balance excursion ($t(8) = 2.32$; $p = 0.049$), and less variability of spatial COP ($t(8) = 2.79$; $p = 0.024$) and COP velocity ($t(8) = 2.31$; $p = 0.05$)

TABLE 1 COP RELATED VARIABLES FOR NO-FF GROUP AND FF-GROUP

| COP variable | FF Group Mean (± SD) | No-FF Group Mean ± SD |
|---|---|---|
| Total Excursion (mm) | 110483.58 (±12038.74) | 141902.65 (±27750.36) |
| RMS COP (mm) | 57.94 (±5.97) | 79.88 (±16.55) |
| RMS COP Velocity (mm/s) | 148.54 (±17.35) | 189.40 (35.58) |

Table 1 represents averaged group COP-related variables. The application of an assistive force field improved standing balance control, as measured indicated by a decrease in excursion and spatial and velocity variability of the COP. RMS = Root Mean Square. COP = Center of Pressure. SD = Standard Deviation. FF = Force field.

### C. Muscle Synergy Analysis

#### 1) Temporal-spatial features of APR: Muscle Tuning

Overall, the muscle tuning curves displayed that the activation level of certain muscles were different between groups but not sensitive to the directionality of the trunk perturbations (Fig. 4). However, the CNS fine-tuned APR differently in the presence of pelvic assistive-force fields (Fig. 5).

*Tibialis Anterior (TA)*

We found a main group effect on TA ($F(1,32) = 4.73$, $p = 0.037$, partial $\eta^2 = 13\%$). The No-FF group (Mean TA activation = 0.022 ±SE = 0.003) showed greater TA activation than the FF group (Mean TA activation = 0.011 ±SE = 0.004) across APR1, APR2 and APR3.

*Lateral Gastrocnemius (LG)*

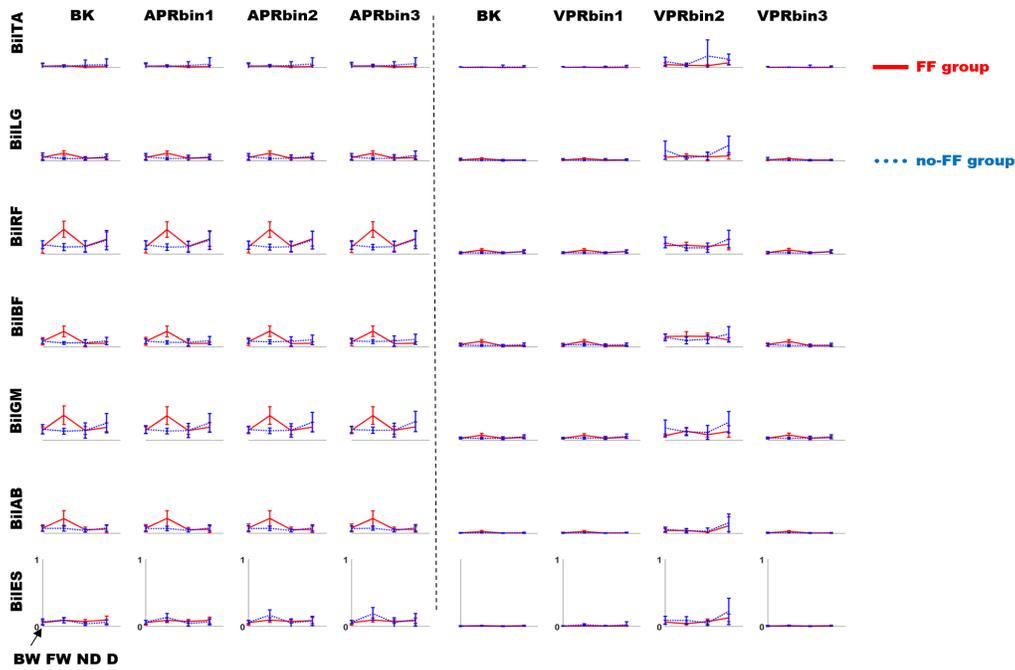

Figure 4 Tuning curves of bilateral postural muscles across perturbation directions. Note how muscles activate differently when the force field was applied. Only LG had a similar activation with and without postural FF. Bil = bilateral, TA = tibialis anterior, LG = lateral gastrocnemius, RF = rectus femoris, BF = biceps femoris, GM = gluteus medius, ABD = rectus abdominis, and ES = erector spinae. BW = Backwards, FW = Forwards, ND = Nondominant, D = Dominant.

We found a Group X Bin interaction effect on LG ($F(6,64) = 2.01$, $p = 0.002$, partial $\eta^2 = 11\%$). In the FF group, the highest level of LG activity was found during APR1 (Mean LG activation during APR1 = 0.045 ±SE = 0.005) compared to APR3 (Mean LG activation during APR3 = 0.042 ±SE = 0.005, $p = 0.009$).

*Rectus Femoris (RF)*

We found a main group effect on RF ($F(1,32) = 4.97$, $p = 0.033$, partial $\eta^2 = 13\%$). Subjects in the FF group demonstrated significantly greater RF activation across APR1, APR2 and APR3 (Mean RF activation = 0.125 ±SE = 0.014) compared to the no-FF group (Mean RF activation = 0.084 ±SE = 0.012).

*Biceps Femoris (BF)*

We found a Group X Bin interaction effect on BF ($F(1.28,41.04) = 8.73$, $p = 0.003$, partial $\eta^2 = 21\%$). In the no-FF group, the APR3 (Mean BF activation during APR3 = 0.099 ±SE = 0.014) was significantly greater than APR1 and APR2 (Mean BF activation during APR1 = 0.078 ±SE = 0.012, $p < 0.001$; and Mean BF activation during APR2 = 0.082 ±SE = 0.012, $p = 0.001$).

*Rectus Abdominis (ABD)*

No significant group differences were found for ABD ($F(1,32) = 3.54$, $p = 0.069$, partial $\eta^2 = 1\%$).

*Erector Espinae (ES)*

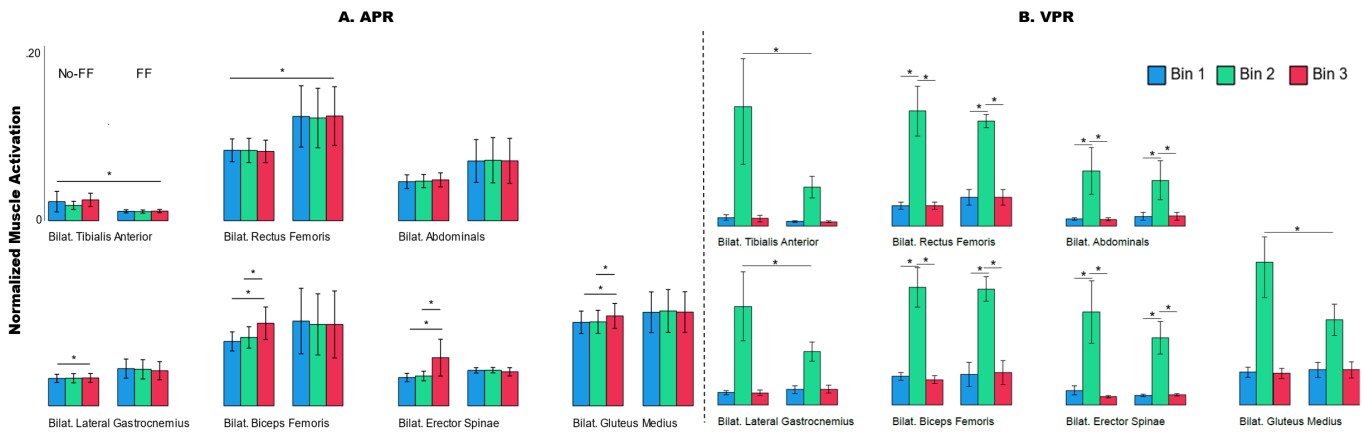

Fig. 5 represents normalized postural automatic (A) and volitional (B) muscle responses across trials, subjects and force field groups (± 2SE). A) The no-FF group showed greater level of activation of TA (a distal muscle related to ankle strategy); whereas the FF group displayed greater muscle activation of RF. In the no-FF group, most of the muscles showed the greatest level of activity within APR3. B) The no-FF group showed greater volitional activity during the ball task in TA, LG and GM than the FF group. * = $p < 0.05$. FF = Force field. Bilat = Bilateral.

We found a marginal Group X Bin interaction effect on ES ($F (1.02, 32.60) = 4.06$, $p = 0.052$, partial $\eta^2 = 11\%$). In the no-FF group, the APR3 (Mean ES activation during APR3 = 0.058 ±SE = 0.009) was significantly greater than APR1 and APR2 (Mean ES activation during APR1 = 0.034 ±SE = 0.002, $p = 0.015$; and Mean ES activation during APR2 = 0.036 ±SE = 0.003, $p = 0.021$).

*Gluteus Medius (GM)*

We found a Group X Bin interaction effect on GM ($F (1.49, 47.60) = 6.33$, $p = 0.007$, partial $\eta^2 = 17\%$). In the no-FF group, APR3 (Mean GM activation during APR3 = 0.108 ±SE = 0.009) showed a significant activation increase than APR1 and APR2 (Mean GM activation during APR1 = 0.101 ±SE = 0.009, $p = 0.001$; and Mean GM activation during APR2 = 0.101 ±SE = 0.009, $p = 0.002$).

2) *Temporal-spatial features of VPR: Muscle Tuning* (Fig. 5)

*Tibialis Anterior (TA)*

We found an interaction group X bin effect on TA ($F (1.009, 32.12) = 4.57$, $p = 0.040$, partial $\eta^2 = 13\%$) between FF and no-FF groups across VPR1, VPR2 and VPR3. The No-FF group showed greater TA activation than the FF group during VPR1 (Mean No-FF = 0.011 ± SE = 0.001 and Mean FF = 0.006 ± SE =0.002, $p = 0.016$) and VPR2 (Mean No-FF = 0.150 ± SE = 0.029 and Mean FF = 0.049 ± SE = 0.036, $p = 0.038$). Only the No-FF group displayed significantly greater TA during VPR2 (Mean no-FF = 0.150 ± SE = 0.029, $p < 0.001$) than VPR1 (Mean no-FF = 0.011 ± SE = 0.001, $p < 0.001$).

*Lateral Gastrocnemius (LG)*

We found an interaction group X bin effect on LG ($F (1.009, 32.279) = 5.81$, $p = 0.022$, partial $\eta^2 = 16\%$) between FF and no-FF groups in VPR2. No-FF group showed greater LG activity in VPR2 compared to VPR1 (Mean No-FF = 0.015 ± SE = 0.002, $p < 0.001$) and VPR3 (Mean No-FF = 0.014 ± SE = 0.002, $p < 0.001$).

*Rectus Femoris (RF)*

We found a main group effect on RF ($F (1.02, 32.47) = 127.74$, $p < 0.001$, partial $\eta^2 = 80\%$). Subjects in both the FF and no-FF groups demonstrated a significant increase of RF activity in VPR2 (mean bin1 = 0.031 ± 0.003, mean bin2 = 0.138 ± 0.010, and mean VPR3 = 0.031 ± 0.002).

*Biceps Femoris (BF)*

Similar to RF, we found main group effect on BF ($F (1.02, 32.47) = 127.74$, $p < 0.001$, partial $\eta^2 = 80\%$). Subjects in both the FF and no-FF groups demonstrated a significant increase of BF activity in VPR2 (mean VPR1 = 0.037 ± 0.004, mean VPR2 = 0.146 ± 0.008, and mean VPR3 = 0.036 ± 0.004, $p < 0.001$).

*Gluteus Medius (GM)*

We found an interaction group X bin effect on GM ($F (1.02, 32.45) = 7.82$, $p = 0.008$, partial $\eta^2 = 20\%$) between FF and no-FF groups in VPR2. No-FF group showed greater GM activation than the FF group (Mean No-FF VPR2 = 0.179 ± SE = 0.019 and Mean FF VPR2 = 0.106 ± SE = 0.024, $p = 0.024$). We found that No-FF group had significantly greater GM during VPR2 than VPR1 (Mean no-FF = 0.041 ± SE = 0.004, $p < 0.001$) and VPR3 (Mean no-FF = 0.039 ± SE = 0.004, $p < 0.001$). The FF group also showed greater GM activity during VPR2 than VPR1 or VPR3, which showed similar levels of muscle activation (Mean FF VPR1 = 0.044 ± SE =0.005 and Mean FF VPR3 = 0.044 ± SE = 0.004, $p < 0.001$).

*Rectus Abdominis (ABD)*

We found a main group effect on ABD ($F (1.00, 32.02) = 25.04$, $p < 0.001$, partial $\eta^2 = 44\%$). Subjects in both the FF and no-FF groups demonstrated a significant increase of RF activity in VPR2 (mean VPR1 = 0.012 ± 0.001, mean VPR2 = 0.066 ± 0.011, and mean VPR3 = 0.011 ± 0.001).

*Erector Espinae (ES)*

We found a main group effect on ES ($F (1.02, 32.58) = 45.85$, $p < 0.001$, partial $\eta^2 = 59\%$). Subjects in both the FF and no-FF groups demonstrated a significant increase of ES activity in VPR2 (mean VPR1 = 0.015 ± 0.002, mean VPR2 = 0.100 ± 0.013, and mean VPR3 = 0.011 ± 0.001).

D. *Postural Synergy Composition during APR and VPR*

The application of a pelvic assistive-force field modulated the composition of postural muscle synergies during APR and VPR (Fig. 6). The synergy module ($W_i$) involved the activation of several muscles that were characterized by diverse anatomical locations between groups. Moreover, muscle synergies were more prevalent on specific perturbation directions and during specific bins ($C_i$).

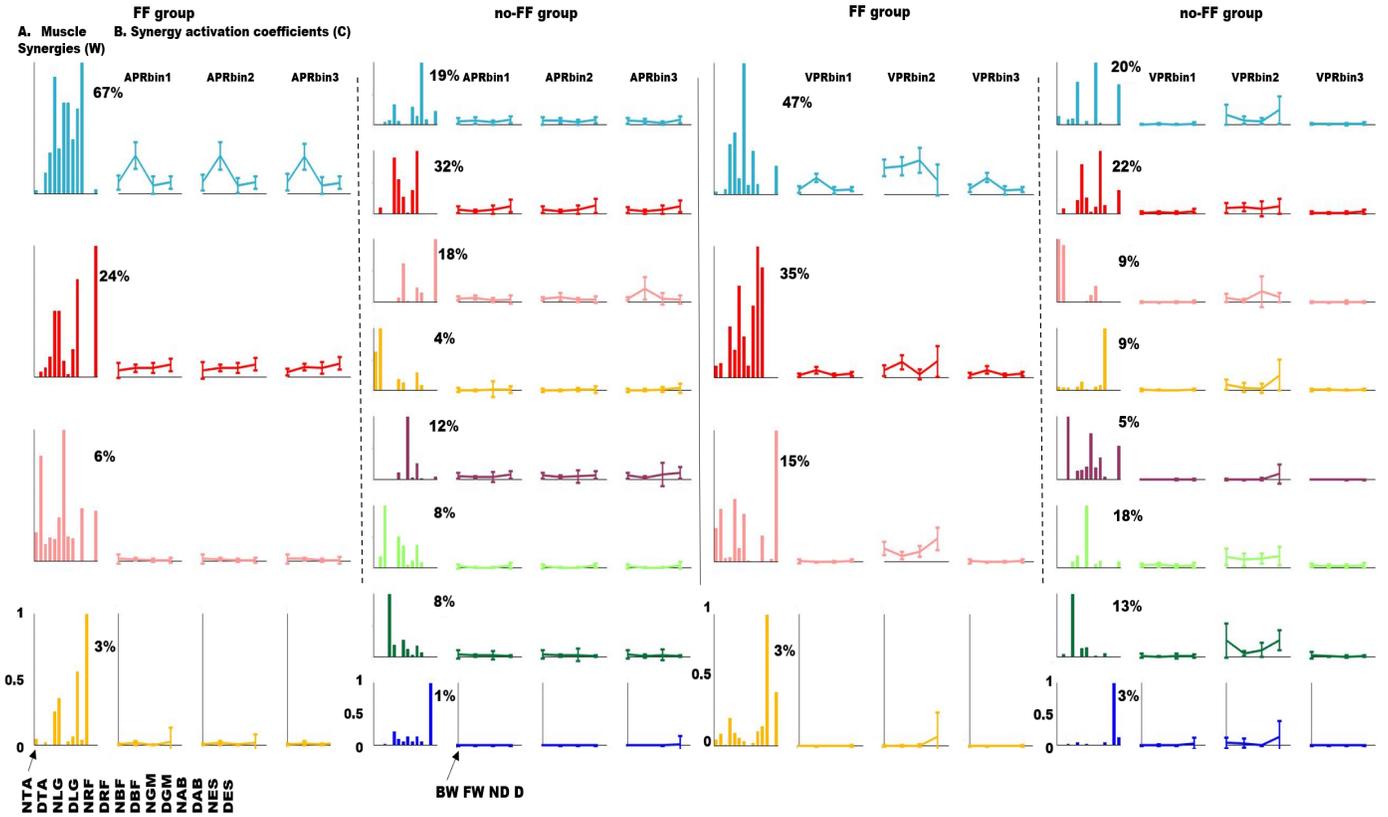

Fig. 6 Muscle synergy vectors and synergy activation coefficients for FF (left panel) and no-FF (right panel) groups. A. muscle synergy vectors, Wi, extracted from EMG data during quiet stance and three APR time bins. Each bar represents the relative level of activation of each muscle within the synergy. B: activation coefficients, Ci, for each of the synergies during each time bin in multiple perturbation directions. NTA =Non-dominant Tibialis Anterior; DTA = Dominant Tibialis Anterior Percentages indicate the amount of total data variability accounted for by each component.

In the FF group, the subjects demonstrated that a total of 4 synergies could optimally reconstruct the background (Bin0) and the three 75ms-bins of each muscle across trials, subjects, and between the two FF conditions—as determined by a VAF = 89% for APR and VAF = 90.31% for VPR. The no-FF group however, required a total of 8 muscle synergies to reach a VAF = 89% and VAF = 91.97% during APR and VPR, respectively.

1) *Functional Organization of Postural Synergies: APR.*

All muscles within a singular synergy, regardless of their level of activation, should play a role in the control of postural adjustments. Nonetheless, if we consider postural muscles highly involved within the synergy ($\geq 0.5$), we can note that subjects in the FF group displayed synergies that predominantly included proximal muscles–which are likely related to a hip strategy. This observation is reflected in muscle synergy *W1* (VAF = 67%), where the biarticular muscles RF (hip flexion) and BF (hip extension) in conjunction with GM (stabilizer) and ABD to control reactive postural adjustments. In the no-FF group, however, these hip-related muscles were distributed across several muscle synergies: synergy *W1* (VAF = 18%) for ABD, synergy *W2* (VAF = 31%) for RF and GM, and synergy *W4* (VAF = 12%) for BF. On the contrary, the no-FF group showed different synergies that mainly recruited distal muscles, which are more likely related to an ankle postural control strategy: synergy *W5* (VAF = 4%) for TA, synergy *W6* (VAF = 8%) for LG, and synergy *W7* (VAF = 8%) for LG. Nonetheless, the FF group only showed synergy *W3* (VAF = 6%) with TA, as a distal muscle potentially involved in an ankle strategy (Fig. 6).

2) *Functional Organization of Postural Synergies: VPR.*

Overall, there was a change in the muscle module composition of the postural synergies during VPR compared to APR in both FF and No-FF groups. Although, this change was perhaps more obvious in the latter group.

In synergy *W1* (VAF = 47%), excluding BF, ABS, GM, and RF were active but less involved (< 0.5) than during APR (Fig. 6). In *W2* (VAF = 20%), GM played a more relevant role than in APR. Also, ES was more involved in synergies *W3* (VAF = 15%) and *W4* (VAF = 3%). In the no-FF group, synergies *W1* (VAF = 20%) and *W2* (VAF = 22%) were more complex and characterized by greater involvement of some hip muscles such as RF and GM. In spite of the presence of hip muscles during VPR, the No-FF group showed a predominance activation of TA within synergy *W3* (VAF = 9%). (Fig. 6)

## IV. DISCUSSION

In this study, we investigated automatic (APR) and volitional (VPR) postural control adjustments in healthy young subjects during postural imbalance and while executing a perceptual multistep motor task in a VR environment (i.e., VR catch-and-throw task). Also, we addressed the modulatory effect on balance via "*assist-as-needed*" force fields at the pelvis mediated by RobUST. We found that subjects receiving assistive-force fields were more successful in the VR catch-and-throw task and demonstrated more proficient postural control during the execution of the task. Furthermore, our outcomes revealed that the CNS orchestrated postural synergies to control standing balance differently during APR and VPR stages with the application of assistive-force fields. Due to the complexity of the postural and reaching task, postural synergies were not direction-specific and as such the perturbation direction did not have a significant impact on muscle tuning.

Previous research has shown that robotic systems that apply force fields can enhance motor performance and induce motor learning [14]. In a previous study, we showed the applicability of RobUST for the dual purpose of delivering direction-specific perturbations and assistive-force fields to supplement the subject's efforts in restoring balance. The main goal of the previously investigated robotic-aided strategy was to promote *in-place* postural control during standing without losing balance (i.e., falls, reaching for support, or taking a step)[16,24]. In more recent studies, we have also demonstrated the potential capability of RobUST, and related motorized cable-driven systems (i.e. trunk-support-trainer), to functionally improve bipedal postural stance and seated postural control during activity-based interventions in people with spinal cord injury and children with cerebral palsy [25,26]. With the present study, we aim at expanding on our previous findings by elucidating how the CNS organizes postural control during a complex VR catch-and-throw task.

*A. Motor behavior & postural control improvements with and without pelvic assistive-force fields*

The attentional, visual perception, motor planning, and high-level of motor coordination required to catch a ball is progressively acquired during motor development. The inherent complexity for catching is such that children do not acquire the motor ability to predict the position of a flying ball and effectively coordinate the control of upper extremities and posture until late childhood—by the age of 10yrs[27–29].

The CNS must also overcome any potential balance disruptions that occur during standing position from the context or secondary to the performance of a motor task such as reaching [9,13]. In our study, we added postural trunk perturbations at the start of the VR catch-and-throw task (see Fig. 2), during the time window that subjects were visually aiming at the flying ball and theoretically preparing the motor plan to catch it (i.e., APR stage). Once the trunk perturbation was delivered, the application of an assistive force field at the pelvis and tailored to the individual's BOS not only enhanced standing balance control but also improved the performance of the VR catch-and-throw task.

*B. Fine-tuning of spatiotemporal muscle activations*

Similar to previous research[13,21], we analyzed multiple time windows during both APR—which includes reactive balance control while visually targeting the ball during the trunk perturbation—and VPR—which includes the resulting preparatory and online postural control strategies during the VR catch-and-throw task[15]. The application of an external assistive-force field to enhance balance control, mainly during the APR stage, changes the CNS organization of postural synergies. We also found that this modulatory force field effect propagated across APR and VPR stages.

Research have found that postural synergies coping with automatic postural responses (<325ms after perturbation) prioritize not only the level of activation but also the muscle onset timing in response to external surface perturbations [12]. Cheung at al (2009) investigated the effect of attaching inertial loads on frog legs during several motor behaviors—kicking, jumping, stepping, and swimming. They found that EMG amplitudes and durations were dependent on the presence of such loads and these electromyographic changes were mediated by sensory signals. These load-related EMG adjustments support that invariant muscle synergies can be modulated to counteract unexpected force demands[30]. Another study applying direction-specific moving platforms in cats found that postural COP-related kinematics and their associated sensory feedback drive the timing and the muscle units that compound postural synergies [10,12]. Similarly, we found that the improvement of COP-related balance kinematics via assistive-force fields results in the activation of different muscles and synergistic postural control. In the No-FF group, participants showed greater activity of TA across APR1, APR2 and APR3; which may functionally be categorized as a key distal muscle for postural control during backward body displacements during an ankle strategy mode of control [16]. The No-FF group particularly showed greater activity of some dorsal postural muscles (BF and ES) and lateral pelvic stabilizer GM—mainly by the end of the APR stage (APR3). In the FF-group however, some muscles such as RF were activated early during APR. These findings support the mechanical effect of the assistive-force field to enhance balance without the need of increased muscle activity to counteract the imposed postural imbalance via perturbative forces [8].

During the VPR, all participants in the FF and No-FF groups activated postural muscles mainly related to a hip control strategy: RF, BF, ABD, and ES. This hip control mode was highly relevant during VPR2, which coincided with the timepoint when these hip-related muscles peaked during the performance of the VR catch-and-throw task. This finding indicates that the postural control of the body during the VR catch-and-throw task mainly rely on a hip control strategy and was developed similarly by FF and No-FF participants. However, during the VPR, the participants in the No-FF group recruited the rest of postural muscles (TA, LG, and GM) with higher level of activity than those included in the FF group; probably because the subjects in the FF group were already stable within the BOS boundaries while performing the VR catch-and-throw task. Furthermore, this finding is in agreement with the better postural performance and greater motor success observed in the FF group during the VR catch-and-throw task. An important biomechanical aspect to notice is that a greater activation of TA and LG in the No-FF can indicate that the participants recruited a mixed hip-ankle postural control strategy. In support of this results, other studies have shown that the CNS can organize postural control within a continuum between ankle and hip control strategies during both perturbed and quiet upright stance [8,31,32]. Studies also show that high-amplitude perturbations can recruit hip control strategies and that trunk movements in standing invoke different modes of postural control—including the knee as an intermediary control strategy [5].

C. *Postural Synergy Composition during APR and VPR*

In spite of the innate character embedded in postural synergies, humans may accomplish goal-oriented actions employing an immense variety of motor strategies—kinematic and kinetic configurations, and spatiotemporal muscle activation patterns [1,12]. Moreover, the mechanical system, motor experience, and surroundings might add some degree of adaptability in what refers to the number and type of muscles involved in a particular postural synergy. Research has shown that synergies are directionally tuned to body COM displacements and muscle units are organized to control the limb forces needed to react against direction-specific postural perturbations [33]. Thus, it is not surprising that in our experiment the application of an assistive-force field modulated the number of postural synergies involved in the control of posture as well as the composition of the synergy vectors ($W_i$). However, we did not find a high-level dependency of direction-specificity in postural synergies. Furthermore, we did not find a strong relationship between muscle tuning curves and synergy coefficient activations per perturbation direction. These findings do not agree with other authors who found that only six synergies could account for specific spatial balance recovery regions during moving platform perturbations [32]. This result is expected however, since our VR catch-and-throw task was complex, highly under volitional control, and characterized by multiple corrective postural control directions secondary to multidirectional perturbative force-fields. Another potential reason is that a different action goal could have been established by the CNS (standing versus standing while planning and performing a complex reaching task); which would demand a different number of postural synergies and muscle units.

Research in reactive postural control has shown that six or fewer postural synergies are enough to control reactive balance during multidirectional platform-based perturbations [8,13,31,32]. In our study, we investigated a different motor task, and we surprisingly found a different number of postural synergies for both FF and No-FF groups. The FF group just recruited a total of four postural synergies to account for the improved postural imbalance while receiving the assistive-force field. Nonetheless, the No-FF group, doubled the number of synergies to eight to control posture during APR and VPR. Our data also revealed that FF participants showed fewer but more complex synergy vectors than No-FF subjects. Therefore, the application of assistive-force fields substantially modified how the CNS constructs and organizes these low-level muscle "*blocks*" to control posture during imbalance conditions or while carrying a complex perceptual-motor reach task. A large number of muscles within each synergy may suggest that the CNS can build up a more effective postural foundation to enable subjects to carry satisfactorily complex reaching actions. Once again, this observation is supported by the fact that FF participants showed greater number of successful catches and throws, and less maximum COP excursions, when RobUST was configured to deliver postural assistive-force fields.

D. *Functional Organization and Activation of Postural Synergies*

As we discussed previously, postural ankle and hip strategies are independent modes of control that can functionally co-exist during postural control, aside from intermediary knee control strategies [15,16]. Most of the previous synergy studies in postural standing have applied perturbations via moving platforms. In a different approach, our RobUST experimental setup uses direction-specific trunk perturbations based on the participant's total body weight percent, as defined by the point of maximum postural instability (i.e., presence of reaching or stepping strategies) [13]. In line with other studies, we found mixed postural strategies in both groups [5,18,33].

Research has shown the existence of anticipatory synergistic muscle control during unfamiliar or challenging self-triggered movements. These feedforward mechanisms occur 100-150ms prior to the initiation of the action and even before the presence of anticipatory postural adjustments [2]. Hence, we believe that in our study, the synergistic muscle control during APR was programmed by sensory feedback. The reason is because the participants had to counteract the trunk perturbation exerted by RobUST while visually tracking the ball in standing position. Nonetheless, our outcomes show that the assistive-force field had an effect on these sensory-based postural synergies. During APR, subjects in the FF group displayed muscles involved in a postural hip strategy (mainly W1); whereas in the No-FF group, the proximal muscles were found within several postural synergies (W1, W2 and W4). No-FF participants presented several synergies involving muscles related to an ankle strategy. Only synergy W3 in the FF group included TA, and to less extent LG distributed across synergies W1, W2, and W3. Overall, there was a change in the muscle module composition of the postural synergies during VPR compared to APR in both FF and No-FF groups. Although, this change was mainly observed in the latter group.

Our results revealed certain changes in the composition of the synergy vectors for both FF and No-FF groups during VPR. For instance, No-FF participants showed postural synergies characterized by greater complexity (W1 and W2) and potentially involved in a hip strategy. In the FF group, the activity of some muscles (ABD in W1 and ES in W2) substantially decreased during VPR. Considering that postural synergies are the neural modules that translate high-level intentions into low-level muscle activations [34], a different anatomical distribution of postural muscles and their level of activation may just represent that that the CNS re-defined two different motor goals during APR (i.e., balance) and VPR (i.e., reaching accuracy). This observation is supported by the lack of correlation of the relative contribution of muscle activations within synergy vectors between APR and VPR stages.

Even though we did not prioritize how the CNS controls reaching in our study, we cannot disregard the high neural complexity involved in the VR catch-and-throw task under investigation in our study groups. Research has shown the involvement of parieto-frontal areas for the control of goal-directed upper extremity movements[34]. A convoluted visual and sensorimotor neural processing would justify once again the need for the CNS to also employ low dimensional synergies to simplify and dictate the mechanical control of the arms via brainstem and spinal cord circuits[2,14,35]. In addition, our participants had to overcome the robotic-mediated postural challenge; which may explain why the No-FF group elicited eight postural synergies to control standing balance while planning and executing the complex VR task.

*E. Study limitations*

When completing complex tasks and maintaining balance, human body needs not only the driving force provided by muscle tissue but also the support and restriction of the skeletal system. The bone structure limits the relative position and range of action of the muscles. Therefore, when studying muscle synergy, not only the optimal dimensionality reduction algorithm of the nervous system, but also the coordination relationship between muscle groups and bones must be considered. Vicon system uses markers to get the precise spatial position and rotation angle of each bone key point of subjects. However, because the infrared positioning equipment of the Vicon system will interfere with the infrared positioning equipment of the VR device, this study did not collect the position information of the bones.

RobUST can provide subjects with controllable and precise force fields on the torso and pelvis levels. However, these force fields are two-dimensional, so they cannot well simulate the three-dimensional interference force field received by the human body in daily life. Based on this limitation, we are improving RobUST to provide a three-dimensional force field.

Another limitation is that we do not have a large sample size. We recruited a total of 10 subjects (5 subjects per group). A larger sample size would likely help to identify any effect of perturbation direction. Besides, in our experiment we didn't analyze the trunk muscles which may be particularly important for the task involving coordination between the legs and arms.

V. CONCLUSION

Our study shows that the application of an assistive-force field significantly modulates how the CNS organizes and fine-tunes postural synergies. The underlying postural synergies of those participants receiving assistive-force fields involved four synergy vectors composed by high activation levels of a minimum of six muscles. This synergy structure was associated with greater motor success and balance control in standing. We also observed a different composition of the synergy vectors between APR (standing during postural imbalance and visually tracking the flying ball) and VPR (executing the VR catch-and-throw task). This different synergy organization was more likely related to the characteristic motor intention of either APR or VPR. The outcomes of the present study address how our activity-based interventions with RobUST may be modulating how the CNS controls posture. Further studies are warranted to investigate this experimental paradigm as a potential postural training in neuromotor populations

ACKNOWLEDGMENT

We would like to thank the contribution of all the participants who collaborated in our study. We also thank Tatiana Luna, MS for her valuable assistance during the experimental setup. We gratefully acknowledge support of the authors from the following grants: NSF IIS-1527087 and New York State SIRB C31290GG, C32238GG.

REFERENCES

1. Ting, L. H. & McKay, J. L. Neuromechanics of muscle synergies for posture and movement. *Current Opinion in Neurobiology* **17**, 622–628 (2007).
2. d'Avella, A. & Lacquaniti, F. Control of reaching movements by muscle synergy combinations. *Frontiers in Computational Neuroscience* **7**, 1–8 (2013).
3. Profeta, V. L. S. & Turvey, M. T. Bernstein's levels of movement construction: A contemporary perspective. *Human Movement Science* **57**, 111–133 (2018).
4. Cheung, V. C. K. *et al.* Stability of muscle synergies for voluntary actions after cortical stroke in humans. *Proceedings of the National Academy of Sciences of the United States of America* **106**, 19563–19568 (2009).
5. Runge, C. F., Shupert, C. L., Horak, F. B. & Zajac, F. E. Ankle and hip postural strategies defined by joint torques. *Gait and Posture* **10**, 161–170 (1999).
6. Shumway-Cook, A. & Woollacott, M. H. Postural Control. in *Motor Control: Translating Research into Clinical Practice* 153–182 (Wolters Kluwer, 2017).
7. Massion, J. Jean Massion_Mov Posture and Equilibrium. *Progress in Neurobiology* **38**, 35–56 (1992).
8. Creath, R., Kiemel, T., Horak, F., Peterka, R. & Jeka, J. A unified view of quiet and perturbed stance: Simultaneous co-existing excitable modes. *Neuroscience Letters* **377**, 75–80 (2005).

9. Cordo, P. J. & Nashner, L. M. Properties of postural adjustments associated with rapid arm movements. *Journal of Neurophysiology* **47**, 287–302 (1982).
10. Girolami, G. L., Shiratori, T. & Aruin, A. S. Anticipatory postural adjustments in children with typical motor development. *Experimental Brain Research* **205**, 153–165 (2010).
11. Hall, L. M., Brauer, S., Horak, F. & Hodges, P. W. With Novel and Familiar Postural Supports Adaptive Changes in Anticipatory Postural Adjustments Adaptive Changes in Anticipatory Postural Adjustments With Novel and Familiar Postural Supports. *J Neurophysiol* **103**, 968–976 (2013).
12. Ting, L. & Macpherson, J. A Limited Set of Muscle Synergies for Force Control During a Postural Task. *Journal of Neurophysiology* **93**, 609–613 (2005).
13. Torres-Oviedo, G. & Ting, L. H. Muscle Synergies Characterizing Human Postural Responses. *Journal of Neurophysiology* **98**, 2144–2156 (2007).
14. D'Avella, A., Portone, A., Fernandez, L. & Lacquaniti, F. Control of fast-reaching movements by muscle synergy combinations. *Journal of Neuroscience* **26**, 7791–7810 (2006).
15. Khan, M. *et al.* Stand Trainer With Applied Forces at the Pelvis and Trunk: Response to Perturbations and Assist-As-Needed Support. *IEEE transactions on neural systems and rehabilitation engineering* **27**, 1855–1864 (2019).
16. Luna, T. D., Santamaria, V., Omofumal, I., Khan, M. I. & Agrawal, S. K. Control Mechanisms in Standing while Simultaneously Receiving Perturbations and Active Assistance from the Robotic Upright Stand Trainer (RobUST). *Proceedings of the IEEE RAS and EMBS International Conference on Biomedical Robotics and Biomechatronics* **2020-Novem**, 396–402 (2020).
17. Krishnamoorthy, V., Goodman, S., Zatsiorsky, V. & Latash, M. L. Muscle synergies during shifts of the center of pressure by standing persons: Identification of muscle modes. *Biological Cybernetics* **89**, 152–161 (2003).
18. Torres-Oviedo, G., Macpherson, J. M. & Ting, L. H. Muscle synergy organization is robust across a variety of postural perturbations. *Journal of Neurophysiology* **96**, 1530–1546 (2006).
19. Torres-Oviedo, G. & Ting, L. H. Muscle synergies characterizing human postural responses. *Journal of Neurophysiology* **98**, 2144–2156 (2007).
20. Chwalisz, K., Diener, E. & Gallagher, D. Autonomic Arousal Feedback and Emotional Experience: Evidence From the Spinal Cord Injured. *Journal of Personality and Social Psychology* **54**, 820–828 (1988).
21. Cheung, V. C. K., D'Avella, A., Tresch, M. C. & Bizzi, E. Central and sensory contributions to the activation and organization of muscle synergies during natural motor behaviors. *Journal of Neuroscience* **25**, 6419–6434 (2005).
22. Lee, D. D. & Seung, H. S. 44565. **401**, 788–791 (2000).
23. Tresch, M. C., Saltiel, P. & Bizzi, E. The construction of movement by the spinal cord. *Nature Neuroscience* **2**, 162–167 (1999).
24. Khan, M. *et al.* Stand Trainer With Applied Forces at the Pelvis and Trunk: Response to Perturbations and Assist-As-Needed Support. *IEEE transactions on neural systems and rehabilitation engineering : a publication of the IEEE Engineering in Medicine and Biology Society* **27**, 1855–1864 (2019).
25. Santamaria V, Luna Tatiana, A. Feasibility and Tolerance of a Robotic Postural Training to Improve Standing in a Person with Ambulatory Spinal Cord Injury. *in submission* (2020).
26. Santamaria, V. *et al.* Promoting Functional and Independent Sitting in Children with Cerebral Palsy Using the Robotic Trunk Support Trainer. *IEEE Transactions on Neural Systems and Rehabilitation Engineering* **28**, 2995–3004 (2020).
27. Bertenthal, B. & von Hofsten, C. Eye, head and trunk control: The foundation for manual development. *Neuroscience and Biobehavioral Reviews* **22**, 515–520 (1998).
28. Huang, M. H. & Brown, S. H. Age differences in the control of postural stability during reaching tasks. *Gait and Posture* **38**, 837–842 (2013).
29. Stamenkovic, A. & Stapley, P. J. Trunk muscles contribute as functional groups to directionality of reaching during stance. *Experimental Brain Research* **234**, 1119–1132 (2016).
30. Cheung, V. C. K., D'Avella, A. & Bizzi, E. Adjustments of motor pattern for load compensation via modulated activations of muscle synergies during natural behaviors. *Journal of Neurophysiology* **101**, 1235–1257 (2009).
31. Alexandrov, A. v., Frolov, A. A. & Massion, J. Biomechanical analysis of movement strategies in human forward trunk bending. I. Modeling. *Biological Cybernetics* **84**, 425–434 (2001).
32. Alexandrov, A., Frolov, A. & Massion, J. Biomechanical analysis of movement strategies in human forward trunk bending. II. Experimental study. *Biol Cybern* **84**, 435–443 (2001).
33. Klous, M., Mikulic, P. & Latash, M. L. Two aspects of feedforward postural control: anticipatory postural adjustments and anticipatory synergy adjustments. *Journal of Neurophysiology* **105**, 2275–2288 (2011).
34. Filimon, F. Human cortical control of hand movements: Parietofrontal networks for reaching, grasping, and pointing. *Neuroscientist* **16**, 388–407 (2010).
35. Sergio, L. E., Hamel-Pâquet, C. & Kalaska, J. F. Motor cortex neural correlates of output kinematics and kinetics during isometric-force and arm-reaching tasks. *Journal of Neurophysiology* **94**, 2353–2378 (2005).